\documentclass[letterpaper, 10 pt, conference]{ieeeconf}
\IEEEoverridecommandlockouts
\usepackage{amsmath,amssymb,amsfonts}
\usepackage{algorithmic}
\usepackage{graphicx}
\usepackage{textcomp}
\usepackage{multirow}
\usepackage{xcolor}
\usepackage{array}
\usepackage{comment}
\usepackage{multirow}
\usepackage[ruled,vlined]{algorithm2e}
\usepackage{setspace}

\newcommand{\red}[1]{\textcolor{black}{#1}}

\usepackage{ctable}

\usepackage{algorithm2e}

\usepackage{subcaption} 

\begin{document}

\title{GRACE: Generating Socially Appropriate Robot Actions Leveraging LLMs and Human Explanations}
\author{Fethiye Irmak Do\u{g}an$^{1}$, Umut Ozyurt$^{2, *}$, Gizem Cinar$^{3, *}$ and Hatice Gunes$^{1}$%
\thanks{$^{1}$ Dept. of Computer Science and Technology, University of Cambridge, UK.  {Corresponding authors: \tt\small \{fid21, hg410\}@cam.ac.uk}. 
}
\thanks{$^{2}$ Dept. of Computer Eng., Middle East Technical University, Turkey.}
\thanks{$^{3}$ Department of Psychology, Bilkent University, Turkey.}
\thanks{* Contributed to this work while undertaking a visiting research studentship at the University of Cambridge's AFAR Lab.}
\thanks{\red{This work has been supported in part by Google under the GIG funding scheme. F. I. Dogan and H. Gunes have been supported by EPSRC/UKRI under grant ref. EP/R030782/1 (ARoEQ). U. Ozyurt and G. Cinar have been supported by the Erasmus Traineeship program.} \red{For open access purposes, the authors have applied a Creative Commons Attribution (CC BY) licence.}}
}


\maketitle
\begin{abstract}


When operating in human environments, robots need to handle complex tasks while both adhering to social norms and accommodating individual preferences. For instance, based on common sense knowledge, a household robot can predict that it should avoid vacuuming during a social gathering, but it may still be uncertain whether it should vacuum before or after having guests. In such cases, integrating common-sense knowledge with human preferences, often conveyed through human explanations, is fundamental yet a challenge for existing systems. In this paper, we introduce GRACE, a novel approach addressing this while generating socially appropriate robot actions. \red{GRACE leverages common sense knowledge from LLMs,  and it integrates this knowledge with human explanations through a generative network}. The bidirectional structure of GRACE enables robots to refine and enhance LLM predictions by utilizing human explanations and makes robots capable of generating such explanations for human-specified actions. \red{Our evaluations show that} integrating human explanations boosts GRACE's performance, where it outperforms several baselines and provides sensible explanations.
\end{abstract}
\section{Introduction}
\vspace{-0.5em}
Robots are expected to operate in challenging real-world environments by learning complex tasks without violating social norms and human preferences. For instance, imagine a robot that is expected to perform household tasks, such as vacuum cleaning or carrying furniture. With the aid of common sense knowledge, the robot can predict that it should not perform these tasks when people are sleeping or when guests are around. However, even with such knowledge, this decision-making process gets complicated in the real world.
For instance, in a home with small children, one person may emphasize safety, preferring the robot to only carry lighter items, while another might prioritize efficiency, expecting the robot to carry multiple heavy objects at once to speed up tasks. \red{In such cases, if the robot could effectively gather insights and explanations about the factors that it should base its decisions on, such as whether to prioritize efficiency, comfort, or safety, it can successfully adjust its actions to human preferences by still adhering to social norms.}

Previous research has explored socially aware robot behaviours across various domains, including household environments~\cite{tjomsland2022mind, checker2024federated, churamani2024feature}, navigation scenarios~\cite{10309661, singamaneni2024survey, gao2022evaluation, tsoi2020sean} and service applications~\cite{9889395, umbrico2020holistic, martins2016context}. These studies have formed a critical foundation for generating robotics systems that are socially appropriate, \red{yet to effectively integrate both social norms and user preferences into robotic actions}, we identify the following critical open challenges: \textbf{\textit{Challenge~1:}} How can the robot decide if is it sufficient to rely on common sense knowledge for the appropriateness of its actions and when should it seek additional insights and explanations from humans around to base its decisions on? 
\textbf{\textit{Challenge~2:}}  
How can the robot effectively integrate these explanations into its decision-making process to ensure its actions 
are aligned with human expectations?
\textbf{\textit{Challenge~3:}} How can the robot generate insights or reasoning for human-specified actions, even when detailed explanations are not provided, in a way that implicitly reflects the humans' (i.e., its user's) underlying preferences and motivations? 
Our work addresses these challenges with a novel methodology that integrates robot action appropriateness with human explanations.

\begin{figure}[t]
    \centering
    \includegraphics[width=0.95\linewidth]{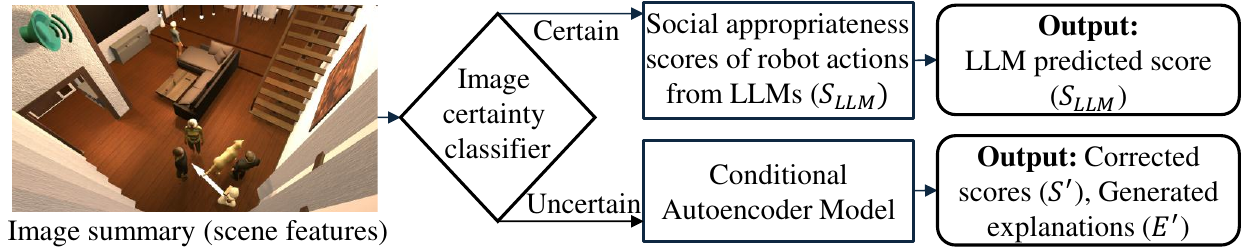}
    \caption{\red{Flowchart of the proposed GRACE system.}}
    \label{fig:Flow-chart}
    \vspace{-2.4em}
\end{figure}

In this paper, we present the so-called system `GRACE' for its ability to generate socially appropriate robot actions.  GRACE is a complete system to generate social appropriateness scores for varying robot actions (such as cleaning, serving food, vacuuming, etc.) in household setups by leveraging common sense knowledge via Large Language Models (LLMs) and integrating this knowledge with human explanations through a generative deep neural network architecture. 
Specifically, GRACE first uses machine learning classifiers to assess the uncertainty of scenes based on human agreement, and for scenes classified as `certain', it directly outputs LLM appropriateness predictions, hence \textit{\textbf{addressing Challenge~1}}.
Critically, for `uncertain' cases, the conditional autoencoder structure of GRACE enhances LLM predictions by integrating them with human explanations \red{(e.g., `there is a child next to the robot', `ensure safety')}, thus \textit{\textbf{addressing Challenge~2}}. Additionally, the bidirectional structure of the GRACE autoencoder enables it to provide coherent explanations for human-annotated scores, thereby \textit{\textbf{addressing Challenge 3}}.  Through extensive evaluations, we demonstrate that GRACE effectively captures the relationship between human scores and their explanations, outperforms several baselines, and \red{boosts performance by leveraging human explanations to predict the appropriateness of robot actions.}

\section{Related Work}
\vspace{-0.5em}
Generating socially aware robot actions has been a critical focus of human-robot interaction (HRI) research, with various studies examining how robots can operate in diverse environments while adhering to social norms and human preferences. These studies dominantly focused on social awareness in navigation~\cite{10309661, singamaneni2024survey, gao2022evaluation, tsoi2020sean, 10161261, 10610505, bacchin2024preference,10610534}, while also contributing to a broader understanding of how robots can align their actions with human expectations across different contexts, such as for service robots~\cite{9889395, umbrico2020holistic, martins2016context} and telepresence systems~\cite{lin2023less, almeida2022telepresence, bacchin2024preference}.
In household environments, Tjomsland et al.\cite{tjomsland2022mind} proposed a continual learning approach using a Bayesian Network to predict human-specified appropriateness scores and their variances, while Churamani et al.~\cite{churamani2024feature} applied federated continual learning to generate socially appropriate robot actions. Although these works provide a strong foundation for our research, they do not account for individual human preferences; instead, they predicted an average score for each scene without incorporating human explanations, which can offer valuable insights into human preferences. 

While enhancing robots' social awareness capabilities, LLMs can serve as an important source to gather common sense knowledge. Accordingly, recent HRI research has utilized LLMs' reasoning abilities in various settings, including assistive applications~\cite{spitale2023vita, spitale2024appropriateness, shi2024can}, service robots~\cite{wang2023wall, wu2023tidybot, ahn2022can}, and ambiguity recognition~\cite{park2023clara, knowno2023}. Former work also deployed LLMs for personalization and preference-based learning in social navigation and mobile manipulation scenarios~\cite{wu2023tidybot, 10.1145/3610977.3634970}. More closely related to our work, Bowen and Harold~\cite{zhang2023large}  utilized LLMs in a zero-shot manner to determine the social appropriateness of robot actions, and they have demonstrated the LLMs' effectiveness in such a task. GRACE adopts this approach to obtain LLM-based predictions, but unlike previous work, we integrate these predictions with human explanations.

While LLMs can offer a critical basis for socially aware robot behaviour, integrating explanations is essential to ensure accountability~\cite{barredoarrieta2020ExplainableArtificialIntelligence}, trust~\cite{siau2018building, edmonds2019tale}, and the efficiency of HRI~\cite{sridharan2019towards, setchi2020explainable, 10.1145/3434073.3444657}.
Despite its importance, this area of research presents significant challenges due to the black-box nature of LLMs~\cite{wu2024usable}, and GRACE makes a timely and novel contribution in this regard.
\textbf{\textit{To our knowledge, GRACE is the first full system to establish a bidirectional approach that integrates LLM-based reasoning with human explanations not only to generate socially appropriate robot actions but also to provide coherent robot explanations.}}

\section{Datasets and Labels}
\vspace{-0.5em}

\subsection{MannersDB and MannerDB+}
\vspace{-0.3em}
To generate socially appropriate robot actions, we utilized two datasets: MANNERSDB~\cite{tjomsland2022mind} (as in~\cite{zhang2023large}) and MANNERSDB+. MANNERSDB contains 750 virtual home settings with a Pepper robot and 11,050 human annotations of these scenes that scored the social appropriateness of eight robot actions (e.g., `\textit{vacuuming}', `\textit{carrying warm food}', etc.) on a scale of 1-5, along with explanations for these scores. MANNERSDB+ extends this dataset to three robots, PR2, Nao, and Pepper, across 1,000 scenes per robot. It includes social appropriateness scores for nine actions, collected from 9,238 human annotators (humans, henceforth), along with their explanations -- see Figure~\ref{fig:explanations} for the complete list of robot actions and example images. In both datasets, each scene includes 29 distinctive features, such as the number of people, their orientation, and whether the music is playing in the background. \red{These scene attributes, annotated scores, and explanations make these datasets highly suitable for evaluating the objectives and capabilities of GRACE.}

 
\subsection{Categorization and Labels of Human Explanations}
\label{categorisation}
\vspace{-0.3em}

\red{To integrate human explanations while generating socially appropriate robot actions, we categorized and labelled human explanations in MannersDB+ (since it has more scenes than MannersDB) by first identifying prominent explanation categories within data and then labelling each sample based on these categories using an LLM.} First, a human evaluator (female, 30 years old, third author of the paper) randomly selected 100 samples from the MannersDB+ dataset and decided on the most recurring explanation categories within these random samples. The evaluator identified nine main categories (e.g., \textit{human state}) and three subcategories based on two opposing (e.g., \textit{available:} 1, and \textit{busy:} -1) and one neutral component (\textit{neutral:} 0)\footnote{\scriptsize Identified explanation categories: \textit{human state (busy, available, neutral)},  \textit{human direction (facing the robot, not facing the robot, neutral)}, \textit{safety (safe, dangerous, neutral)}, \textit{robot direction (facing humans, not facing humans, neutral)}, \textit{working area (big, small, neutral)}, \textit{robot capability (capable, incapable, neutral)}, \textit{robot proximity (close, far, neutral)}, \textit{crowd (crowded, empty neutral)}, \textit{environmental noise (noisy, silent, neutral)}.}.

For each category, we labelled human explanations in MannersDB+ by querying GPT-4o-mini  separately. We applied a two-shot prompting technique in each category by providing examples from the opposing components. 
To ensure the correctness of LLM labels, the same human evaluator selected another random 200 samples, manually labelled them based on pre-defined categories, and compared them with LLM-generated ones, which resulted in a high match (95.28\%) and confirmed the precision of the LLM-generated labels. 
After obtaining the LLM labels, we had to discard \textit{human direction} and \textit{environmental noise} categories due to inadequate samples. Overall, for each explanation, we obtained a seven-dimensional vector, and each of these dimensions could take three different values (-1, 0, or 1), which we later normalized each dimension between 0 and 1.


\section{METHODOLOGY}
\vspace{-0.5em}
In this section, we conceptualize the novel idea of generating socially appropriate robot actions by leveraging large language models (LLMs) and human annotator explanations. 
For a given scene, we first assess the uncertainty of the scene, i.e., if human annotators would agree on the scores for robot action appropriateness. To achieve this, we obtain binary labels with unsupervised clustering and train ML classifiers. This classification aims to identify when LLM predictions would be enough to predict robot action appropriateness (certain scenes, i.e., humans would agree on their scores) or when LLM predictions should be improved with explanations (uncertain scenes, i.e., humans would have disagreements). For uncertain situations, we suggest an autoencoder-based generative model that leverages human explanations to enhance the generation of socially appropriate robot actions, as well as endow the robot with the capability to generate explanations when necessary \red{(overall pipeline in Figure~\ref{fig:Flow-chart}).}

\subsection{Scene Clustering and Uncertainty Classification}
\vspace{-0.3em}

To determine the certainty of scenes, we rely on human annotator agreement, i.e., classifying scenes based on when human annotators are most likely to give similar scores for the appropriateness of robot actions (certain scenes), or they are likely to give different ones (uncertain scenes). We formulate this task as a binary class classification problem. \red{We first obtain the weak binary labels for each scene (showing if a scene is certain or uncertain) with unsupervised clustering and then employ various classifiers to predict uncertainties.}



To obtain the binary labels for each scene $x_{i} \in X$, we consider the variance of different human scores for the appropriateness of each action $a_{j} \in A$ (e.g., \textit{`vacuuming'}, \textit{`starting a conversation'}, \textit{`serving warm food'}, etc.). More specifically, we compute the set of variances: $V_{x_i} = \{\sigma^2(s_{a_1}^{x_i}), \sigma^2(s_{a_2}^{x_i}), \dots, \sigma^2(s_{a_n}^{x_i})\}$,
where $\sigma^2(s_{a_j}^{x_i})$ is the variance of human scores $s_{a_j}^{x_i} \in S_{human}$ for action $a_j$ in scene $x_i$, and $n$ is the total number of different actions in the set $A$. The set $V_{x_i} \in V_X$ represents the variances for all actions in scene $x_i$, intuitively reflecting human annotators' level of agreement for that scene. We then input the variance feature set $V_X$ into the K-means++ clustering algorithm to group the scenes into two clusters, resulting in two centroids. The centroid with lower feature values (i.e., lower variance) is set as the \textbf{certain} cluster centroid, and the other one is identified as \textbf{uncertain}. Finally, each scene $x_i$ is assigned a binary label $y_{x_i} \in \{0, 1\}$, where $y_{x_i} = 1$ if $V_{x_i}$ is closer to the uncertain centroid based on its Euclidean distance, and $y_{x_i} = 0$ otherwise.


After generating the weak labels for each scene $x_i$, these labels serve as the target variable $y_{x_i} \in \{0, 1\}$ in a binary classification task to predict scene uncertainties. The input space of the classification tasks is obtained from scene features, i.e., for each scene, $x_i$, the set of scene features is $f_{x_i} = \{f^1_{x_{i}}, f^2_{x_{i}}, \dots, f^p_{x_{i}}\}$,
where $f_{x_i}$ includes $p$ number of descriptive attributes such as the number of people in the scene, their relative directions, if the music is playing in the background, etc. For the prediction process, we employ varying machine learning models: \red{logistic regression (LR), random forest (RF), support vector machine (SVM), Multi-layer perceptron (MLP), gradient boosting (GB), and k-nearest neighbour (KNN) classifiers}. Given a trained classifier $g$ and its model parameters $\theta$, the classifier predicts the label $\hat{y}_{x_i} = g( \{f^1_{x_{i}}, f^2_{x_{i}}, \dots, f^p_{x_{i}}\}; \theta)$ for each scene. The training details are provided in Section~\ref{uncertain_class_details}.

\subsection{Action Appropriateness using LLMs}
\vspace{-0.3em}

To generate the scores for the social appropriateness of robot actions with LLMs, we first obtain the textual descriptions of a scene $x_i$ from its feature set $f_{x_i}$. To generate the prompt including these features, we follow the approach and implementation\footnote{\scriptsize https://github.com/clear-nus/llm-human-model} of Bowen and Harold~\cite{zhang2023large} on MannersDB Data. 
In this implementation, each scene is textually summarized based on its attributes, and LLMs are queried separately for each action $a_j$. While querying, we ensure that LLMs generate the top $k$ most probable outputs, where $k$ corresponds to the number of possible score (appropriateness level) choices. Then, the overall score for an action $a_j$ in scene $x_i$ is calculated as the weighted sum of LLM outputs by multiplying each LLM-generated score with its probabilities and then taking their sum. This approach yields an LLM-generated score $\hat{s}_{a_j}^{x_i} \in S_{LLM}$ for each action $a_j$ in a given scene $x_i$. The implementation details are provided in Section~\ref{LLM_details}.

\begin{figure}
    \centering
    \includegraphics[width=1\linewidth]{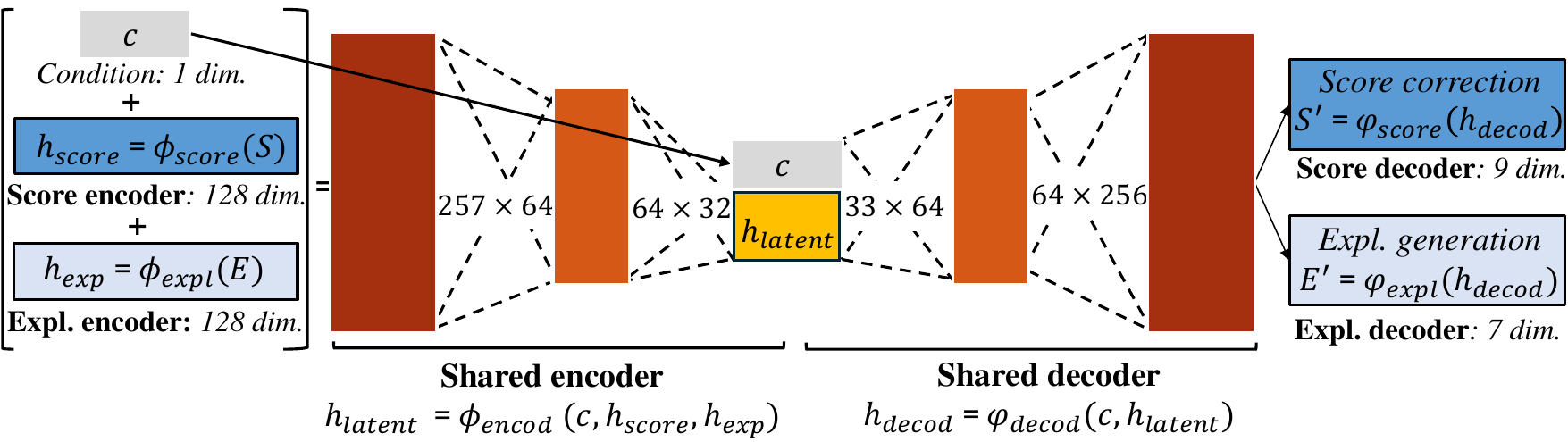}
    \caption{The network structure of the GRACE autoencoder.
   }
    \label{autoencoder}
    \vspace{-1.5em}
\end{figure}

\subsection{Leveraging Explanations for Action Appropriateness}
\vspace{-0.3em}

Former sections focus on how to decide if a given scene is `certain' and how to use LLMs to predict the score values for the social appropriateness of robot actions when there are `no uncertainties'. On the other hand, in this section, we focus on handling `uncertain' scenes (where humans would have disagreements) and propose a generative model that leverages LLM outputs and human explanations to enhance the predictions. The structure of this model not only enables the improvement of the LLM score predictions with human explanations but is also capable of generating explanations for given human scores. To achieve this, we employ a conditional autoencoder structure designed to handle both score correction and explanation generation tasks.


First, the input of the GRACE autoencoder includes both LLM-predicted scores and human scores, depending on the condition, as well as human explanations or noise. The input $I$ of the network is defined as:\vspace{-0.3em}
\begin{equation}
I =
\begin{cases}
(c_0, S_{LLM}, E_{human}), & \text{ score correction}, \\
(c_1, S_{human}, E_{noise}), & \text{\red{ expl. generation}},
\end{cases} 
\vspace{-0.3em}
\end{equation}%
where $c_0$ is the condition for score correction, and $c_1$ is for explanation generation. Additionally, $S_{LLM}$ shows the vector of LLM-predicted scores, whereas $S_{human}$ is the vector of human scores, which both have $n$ dimensions for different robot actions. Lastly, $E_{human}$ represents the human explanations (one-hot encoded and has $m$ dimensions denoting the different categories of human explanations, e.g., the robot's proximity dimension is one if the human's explanation states that the robot is far away from people, it is zero if the robot is too close to people, and 0.5 otherwise), and $E_{noise}$ is random explanations generated from noise. The target value for both conditional inputs $I_{c_0}$ and $I_{c_1}$ is set as $(S_{human}, E_{human})$, aiming to reconstruct human scores and explanations.

The encoder part processes both scores and explanations and combines them into a shared latent representation. First, the score encoder takes the score vector $S$ (either comes from $S_{LLM}$ or $S_{human}$) and transforms it into a latent vector $h_{\text{score}}$. Similarly, the explanation encoder transforms the explanation vector $E$ (either comes from $E_{human}$ or $E_{noise}$) into another latent vector $h_{\text{expl}}$. Finally, the encoded score and explanation vectors, along with the condition $c$ (either $c_0$ or $c_1$), are concatenated and passed through shared encoder layers:\vspace{-0.5em}
\begin{equation}
    \begin{aligned}
        h_{\text{score}} &= \phi_{\text{score}}(S), \quad
        h_{\text{exp.}} = \phi_{\text{expl}}(E), \\
        &h_{\text{latent}} = \phi_{\text{encod}}(c, h_{\text{score}}, h_{\text{expl}}).
    \end{aligned}
\end{equation}%
In the decoding phase, the shared latent vector $h_{\text{latent}}$ reconstructs both the scores and explanations. First, the shared decoder takes the latent vector $h_{\text{latent}}$ and the condition $c$,  and it produces an intermediate representation, $h_{\text{decod}}$. Then, this shared representation is transformed back into the predicted score vector $S'$ by the score decoder. Similarly, the explanation decoder reconstructs the explanation vector $E'$ from $h_{\text{decod}}$:\vspace{-0.5em}
\begin{equation}
    \begin{aligned}
    &h_{\text{decod}} = \varphi_{\text{decod}}(c, h_{\text{latent}}),\\
        S' &= \varphi_{\text{score}}(h_{\text{decod}}), \quad
        E' = \varphi_{\text{expl}}(h_{\text{decod}}).
    \end{aligned}
\end{equation}%
During training, the autoencoder model uses a combined loss to balance both score prediction and explanation generation:
\begin{equation}
    \mathcal{L} = \alpha \cdot \text{MSE}(S_{human}, S') + (1 - \alpha) \cdot \text{BCE}(E_{human}, E')
\end{equation}%
\noindent where MSE is the mean squared error between the predicted scores $S'$ and the human scores $S_{human}$, and BCE shows the binary cross-entropy between the predicted explanations $E'$ and the human explanations $E_{human}$. Lastly, $\alpha$ represents a weight that balances the contributions of the score and explanation losses. The detailed structure of the autoencoder model is shown in Figure~\ref{autoencoder}, and the training procedure and model parameters are provided in Section~\ref{Generative_training}.

\section{Experiments and Results}
\subsection{Implementation Details, Baselines and Metrics}
\vspace{-0.3em}
\subsubsection{Uncertainty Classification}
\label{uncertain_class_details}


For uncertainty classification, we trained all machine learning models with the scikit-learn library\footnote{\scriptsize https://scikit-learn.org/stable/} on MannersDB and MannersDB+ datasets. While training, we avoided data leakage and utilized a nested cross-validation approach with five inner and five outer loops, where the inner loop was used for the hyperparameter tuning with a randomized search. 
In the experiments with MannersDB and MannersDB+, the total number of robot actions ($n$) was 8 and 9, and the number of descriptive scene attributes ($p$) was 29 and 32, respectively. The three additional attributes in MannersDB+ came from the one-hot encoding of three robots.



The weak label generation with K-means++ led to a class imbalance problem. To mitigate this, the `balanced' weight option was utilized during hyper-parameter tuning, adjusting weights based on class frequencies. For models lacking this feature (MLP, GB, KNN), minority class oversampling was applied. Additionally, we explored bootstrap aggregation (bagging) to further assess performances. Model performances were measured with balanced accuracy (Unweighted Average Recall) to ensure equal weight for both classes. F1 score and Precision were reported as macro averages, calculated by averaging metrics per class without weighting.



In addition to ML models, we obtained a \textbf{human baseline} for scene uncertainty classification. We randomly sampled 100 scenes from MannersDB and MannersDB+ and asked human evaluators if they think people would agree or disagree (e.g., give similar or different scores) on the appropriateness of robot actions. 9 evaluators (4 female, 4 male, one non-binary, avg. age: 24 $\pm$ 5.29) evaluated MannersDB samples, whereas 11 evaluators (4 female, 6 male, one non-binary, avg. age: 24.27 $\pm$ 4.94) completed the task for MannersDB+. This baseline provides a reference for human performances on uncertainty classification, as well as gives an indication of task difficulty.

\subsubsection{LLM Predictions}
\label{LLM_details}



In order to obtain LLM predictions for robot actions' social appropriateness, we experimented with Google/flan-t5-xxl, Llama-3.1 8B-Instruct, Llama-3.1 70B-Instruct, GPT-4o-mini and GPT-4o language models on MannersDB+. Building on the querying method by Bowen and Harold \cite{zhang2023large}, we adjusted the prompts to include the robot type and restricted the models’ responses to only output the appropriateness of the actions. For Llama models, we further enforced this constraint by suppressing the generation of nine words, including `\textit{The}', `\textit{Answer}', and an empty string.

\red{In the experiments, the number of most probable outputs ($k$) was set to five to cover all the options ranging from `very inappropriate' (score 1) to `very appropriate' (score 5)}. We reported the accuracies of LLMs based on Root Mean Square Error (RMSE, lower is better), Pearson Correlation Coefficient (PCC, higher is better), and  Concordance Correlation Coefficient (CCC, higher is better) to measure how closely LLM predictions matched human scores.


\subsubsection{Leveraging Explanations for Action Appropriateness}
\label{Generative_training}

To assess the impacts of leveraging explanations, we evaluated the GRACE autoencoder on MannersDB+, as it offers more scenes than MannersDB (3000 scenes vs 750 scenes). Similar to Section~\ref{uncertain_class_details}, we used a nested cross-validation approach with five inner and five outer loops and avoided data leakage by assigning the samples coming from the same image to the same split (either train, test, or validation). The network parameters were decided with a grid search: The network had 256 and 64 dimensions in the shared embeddings and 32 dimensions in the latent one. In between these layers, ReLU nonlinearity~\cite{NIPS2012_c399862d} is used to address vanishing gradients, and the network was trained with the ADAM optimizer~\cite{kingma2014adam} (beta1: 0.9, beta2: 0.999, and weight decay: 1e-8). The initial learning rate was set as 1e-3, decaying by 0.3 after ten epochs of no improvement. The loss balance parameter $\alpha$ was set to 0.6, the network was trained with a batch size of 32 for 200 epochs, and early-stopping was triggered if the validation accuracies did not improve more than an $\epsilon$ value (set as 0.0001) for 20 epochs. Lastly, the number of different actions ($n$) was nine, whereas the total number of human explanation categories ($m$) was seven, the condition $c_O$ was set as zero, the condition $c_1$  was set as one, and the random explanations $E_{noise}$ were set as the vector of 0.5 values during the experiments.

We conducted experiments with various baselines that follow two distinct approaches for generating socially appropriate robot actions. The first set includes methods that predict the appropriations scores based on scene features~\cite{zhang2023large, tjomsland2022mind}. For Bowen and Harold's LLM-based approach~\cite{zhang2023large}, we obtain the LLM outputs as described before, and for Tjomsland et al.~\cite{tjomsland2022mind}, we implemented the MLP model (shown to perform the best in their paper) with their specifications. The second set of baseline approaches includes correcting LLM scores to better match them with human scores without including explanations. For this, we trained Autoencoders (AE), Variational Autoencoders (VA), and Denoising Autoencoders (DAE) using a Salt and Pepper noise with 0.2 probability. All these models used the same parameters as the GRACE autoencoder. We additionally evaluated the denoising approach in our model by adding the noise to the part that the network aims to reconstruct (e.g., either the score or explanation) based on the given condition. All model performances were evaluated with RMSE, PCC, and CCC metrics by analyzing how well the model-generated scores match with the human ones.




\subsection{Results}
\vspace{-0.3em}
\subsubsection{Uncertainty Classifications}

This section presents uncertainty classification results for MannersDB and MannersDB+ by comparing ML models based on their match with the cluster-based ground truth. For MannersDB, the SVM method achieved the best performances among all models (mean values on the left side of Table~\ref{tab:uncertainty_pred_combined}, with STD values from cross-validation ranging between 0.01 and 0.05). Oversampling improved GB performances but decreased KNN and MLP accuracies. After identifying SVM as the best model, we also applied bagging to SVM, which resulted in a slight performance increase. For MannersDB+, overall accuracies were lower, and all models performed similarly (right side of Table~\ref{tab:uncertainty_pred_combined}, STD between 0.01 and 0.04). \red{For models without built-in class weight balancing (GB, KNN, MLP), oversampling improved balanced accuracy and F1 scores but decreased precision. Bagging did not affect SVM performances. Lastly, the human baseline showed poor performance for both datasets, highlighting task difficulty.}

\begin{table}[t!]
    \caption{Scene uncertainty classification performances on test sets for MannersDB (left) and MannersDB+ (right). Oversampling (Ov) and Bagging (B) results in parenthesis.}
    \centering
    \scriptsize
    \setlength{\tabcolsep}{1pt} 
    \renewcommand{\arraystretch}{1} 
    \begin{tabular}{c|c | c | c||c |c |c}
         & \multicolumn{3}{c|| }{\textbf{MannersDB}} & \multicolumn{3}{c}{\textbf{MannersDB+}} \\
         & Accuracy & F1 & Precision & Accuracy  & F1 & Precision \\ 
         & (balanced) & (macro) & (macro) &  (balanced) & (macro) & (macro) \\\hline
         Human   & 0.50  & 0.49  & 0.50  & 0.45  & 0.44  & 0.45  \\\hline\hline
         LR      & 0.60  & 0.59  & 0.59  & \textbf{0.56}  & \textbf{0.55}  & 0.55  \\ \hline
         RF      & 0.61  & 0.60  & 0.60  & 0.54  & 0.54  & 0.54  \\ \hline
         GB (Ov)     & 0.58 (0.60)  & 0.57 (0.59)  & 0.58 (0.59)  & 0.52 (0.53)  & 0.51 (0.53)  & 0.53 (0.53)  \\ \hline
         KNN (Ov)    & 0.60 (0.59)  & 0.59 (0.58)  & 0.60 (0.59)  & 0.52 (0.53)  & 0.48 (0.52)  & 0.56 (0.53)  \\ \hline
         MLP   (Ov)  & 0.59 (0.58)  & 0.58 (0.57)  & 0.61 (0.57)  & 0.52 (0.54)  & 0.47 (0.53)  & \textbf{0.60} (0.54)  \\ \hline
         SVM   (B)  & 0.63 (\textbf{0.64})  & 0.62 (\textbf{0.63})  & 0.62 (\textbf{0.64})  & 0.55 (0.55)  & \textbf{0.55} (\textbf{0.55})  & 0.55 (0.55)  \\
    \end{tabular}
    \label{tab:uncertainty_pred_combined}
    \vspace{-2.5em}
\end{table}

\subsubsection{Cluster Analysis and LLM Performances}

\begin{table*}[t]
    \setlength{\tabcolsep}{2pt}
    \caption{LLM performances based on clusters. 
    (WD: Whole Data, UC: Uncertain Cluster, CC: Certain Cluster )}
    \centering
    \scriptsize
    \begin{tabular}{c|c|c|c|c|c}
        \centering
        \setlength{\tabcolsep}{3pt}
        & \centering Google/flan-t5-xxl & Llama-3.1 8B-Instruct & Llama-3.1 70B-Instruct & GPT-4o-mini & GPT-4o \\
        & \centering \hspace{1.25em} RMSE \hspace{1.25em} \textbar \hspace{-0.1em}  PCC \hspace{-0.38em} \textbar\ CCC & \hspace{1.25em} RMSE \hspace{1.25em} \textbar \hspace{-0.1em}  PCC \hspace{-0.38em} \textbar\ CCC & \hspace{1.25em} RMSE \hspace{1.25em} \textbar \hspace{-0.1em}  PCC \hspace{-0.38em} \textbar\ CCC & \hspace{1.25em} RMSE \hspace{1.25em} \textbar \hspace{-0.1em}  PCC \hspace{-0.38em} \textbar\ CCC & \hspace{1.25em} RMSE \hspace{1.25em} \textbar \hspace{-0.1em}  PCC \hspace{-0.38em} \textbar\ CCC \\\hline
        WD & \centering 1.30 ($\pm$ 0.46) \textbar\ 0.25 \textbar\ 0.08  & 1.59 ($\pm$ 0.64) \textbar\  \textbf{0.15} \textbar\ 0.03  & 1.37 ($\pm$ 0.52) \textbar\ \textbf{0.20} \textbar\ \textbf{0.05}  & 1.39 ($\pm$ 0.51) \textbar\ 0.26 \textbar\ 0.14  & 1.37 ($\pm$ 0.51) \textbar\ 0.34 \textbar\ 0.23  \\
        UC & \centering 1.48 ($\pm$ 0.47) \textbar\ 0.25 \textbar\ 0.07  & 1.76 ($\pm$ 0.70) \textbar\  \textbf{0.15} \textbar\ 0.03  & 1.56 ($\pm$ 0.54) \textbar\ 0.19 \textbar\ 0.04  & 1.54 ($\pm$ 0.53) \textbar\ 0.23 \textbar\  0.12  & 1.54 ($\pm$ 0.54) \textbar\ 0.32 \textbar\ 0.20 \\
        CC & \centering \textbf{1.21} ($\pm$ 0.43) \textbar\ \textbf{0.26} \textbar\ \textbf{0.09}  & \textbf{1.50} ($\pm$ 0.58) \textbar\ \textbf{0.15} \textbar\ \textbf{0.04}  & \textbf{1.27} ($\pm$ 0.49) \textbar\ \textbf{0.20} \textbar\ \textbf{0.05} & \textbf{1.31} ($\pm$ 0.47) \textbar\ \textbf{0.27} \textbar\ \textbf{0.15}  & \textbf{1.28} ($\pm$ 0.47) \textbar\ \textbf{0.35} \textbar\ \textbf{0.24}  \\
    \end{tabular}
    \label{table:LLM_performances}
    \vspace{-0.5em}
\end{table*}

\begin{table*}[t]
    \caption{Performances of different models and GRACE on the test data. The models that predict the social appropriateness of robot actions using scene features are shown on the left. The models that correct LLM scores are presented on the right. 
    (AE: Autoencoder, VAE: Variational AE, DAE: Denoising AE, GRACE (Expl.) + N: Proposed model with a noising mask).}
    \vspace{-0.5em}
    \centering%
    \scriptsize
    \begin{minipage}[t]{0.25\textwidth}  
        \centering
        \setlength{\tabcolsep}{1.85pt} 
        \begin{tabular}{c|c c c}
        \textit{Approaches using scene}  \\
             \textit{features for score prediction} & RMSE & PCC & CCC \\\hline
            MLP~\cite{tjomsland2022mind} & 1.21  & 0.42 & 0.39 \\
            LLM (Flan-T5)~\cite{zhang2023large} & 1.30  & 0.25 &  0.08\\
            LLM (Llama 70B)~\cite{zhang2023large} & 1.37 & 0.20 &  0.05\\
            LLM (GPT-4o)~\cite{zhang2023large} & 1.37  & 0.34 &  0.23 \\
            \textbf{GRACE} with  GPT-4o & \textbf{1.10}  & \textbf{0.52} & \textbf{0.48} \\
        \end{tabular}
    \end{minipage}%
    \hspace{2.8em} 
    \begin{minipage}[t]{0.7\textwidth}  
        \centering
        \setlength{\tabcolsep}{1.85pt} 
        \scriptsize
        \begin{tabular}{c|ccc|ccc|ccc|ccc|ccc}
            \textit{Models correcting} & \multicolumn{3}{c|}{Flan-T5} & \multicolumn{3}{c|}{Llama 8B} & \multicolumn{3}{c|}{Llama 70B} & \multicolumn{3}{c|}{GPT-4o-mini} & \multicolumn{3}{c}{GPT-4o} \\
            \textit{LLM scores} & RMSE & PCC & CCC & RMSE & PCC & CCC & RMSE & PCC & CCC & RMSE & PCC & CCC & RMSE & PCC & CCC \\
            \hline
            AE (No Expl.)  & 1.22 & 0.45 & 0.42 & 1.28 & 0.34 & 0.32 & 1.23 & 0.44 & 0.40 & 1.22 & 0.46 & 0.42 & 1.21 & 0.47 & 0.43 \\
            VAE (No Expl.)  & 1.23 & 0.43 & 0.40 & 1.23 & 0.43 & 0.40 & 1.23 & 0.43 & 0.40 & 1.22 & 0.46 & 0.42 & 1.22 & 0.46 & 0.43 \\
            DAE (No Expl.) & 1.22 & 0.44 & 0.41 & 1.28 & 0.34 & 0.31 & 1.24 & 0.41 & 0.38 & 1.22 & 0.45 & 0.41 & 1.22 & 0.46 & 0.43 \\
            \textbf{GRACE} (Expl.)  & \textbf{1.08} & \textbf{0.56} & \textbf{0.52} & \textbf{1.09} & 0.55 & \textbf{0.51} & \textbf{1.08} & \textbf{0.56} & \textbf{0.52} & 1.08 & \textbf{0.57} & \textbf{0.52} & 1.09 & \textbf{0.57} & 0.52 \\
            \textbf{GRACE} (Expl.) + N & \textbf{1.08} & \textbf{0.56} & \textbf{0.52} & \textbf{1.09} & \textbf{0.56} & \textbf{0.51} & \textbf{1.08} & \textbf{0.56} & \textbf{0.52} & \textbf{1.07} & 0.56 & \textbf{0.52} & \textbf{1.07} & \textbf{0.57} & \textbf{0.53} \\
        \end{tabular}
    \end{minipage}
    \label{combined_tables}
    \vspace{-2.8em}
\end{table*}

First, we analyzed the entire MannersDB+ and each cluster based on their aleatoric uncertainty.
\red{To compute the uncertainty, as in Jonas et al.~\cite{tjomsland2022mind}, we took the mean of log variances of human scores for each scene, i.e., $log\sigma^2(s_{a_j}^{x_i})$.} Expectedly, the lowest value was obtained for the certain cluster  (-0.59 $\pm$ 0.41), whereas it was higher for the whole data (-0.31 $\pm$ 0.56) and the highest for the uncertain cluster (0.25 $\pm$ 0.36). 

Next, we show LLM prediction performances on the entire MannersDB+ and each cluster in Table~\ref{table:LLM_performances}, 
where we compare different LLMs based on how well their predictions match with human scores. Once again, expectedly, all models showed lower RMSE values and almost always the highest correlation coefficients (PCC and CCC) for samples in the certain cluster, whereas these performances were the worst for the uncertain cluster, and the whole data was in between. Additionally, when we compare the performances of different LLM models, we see that they all reported similar RMSE values (Google/flan-t5-xxl slightly better), \red{whereas the correlation coefficients were substantially higher for GPT-4o, making it the best at capturing the trend in human scores.}

\subsubsection{Action Appropriateness using Explanations}

\begin{figure}[t]
    \centering
    \scriptsize
    \captionsetup[subfigure]{labelformat=empty}
    \setlength{\tabcolsep}{2pt} 
    \renewcommand{\arraystretch}{0.8} 
    
    \centering
    \begin{subfigure}[b]{0.321\linewidth} 
        \centering
        \caption{\textbf{Scene A}}
        \includegraphics[width=\linewidth]{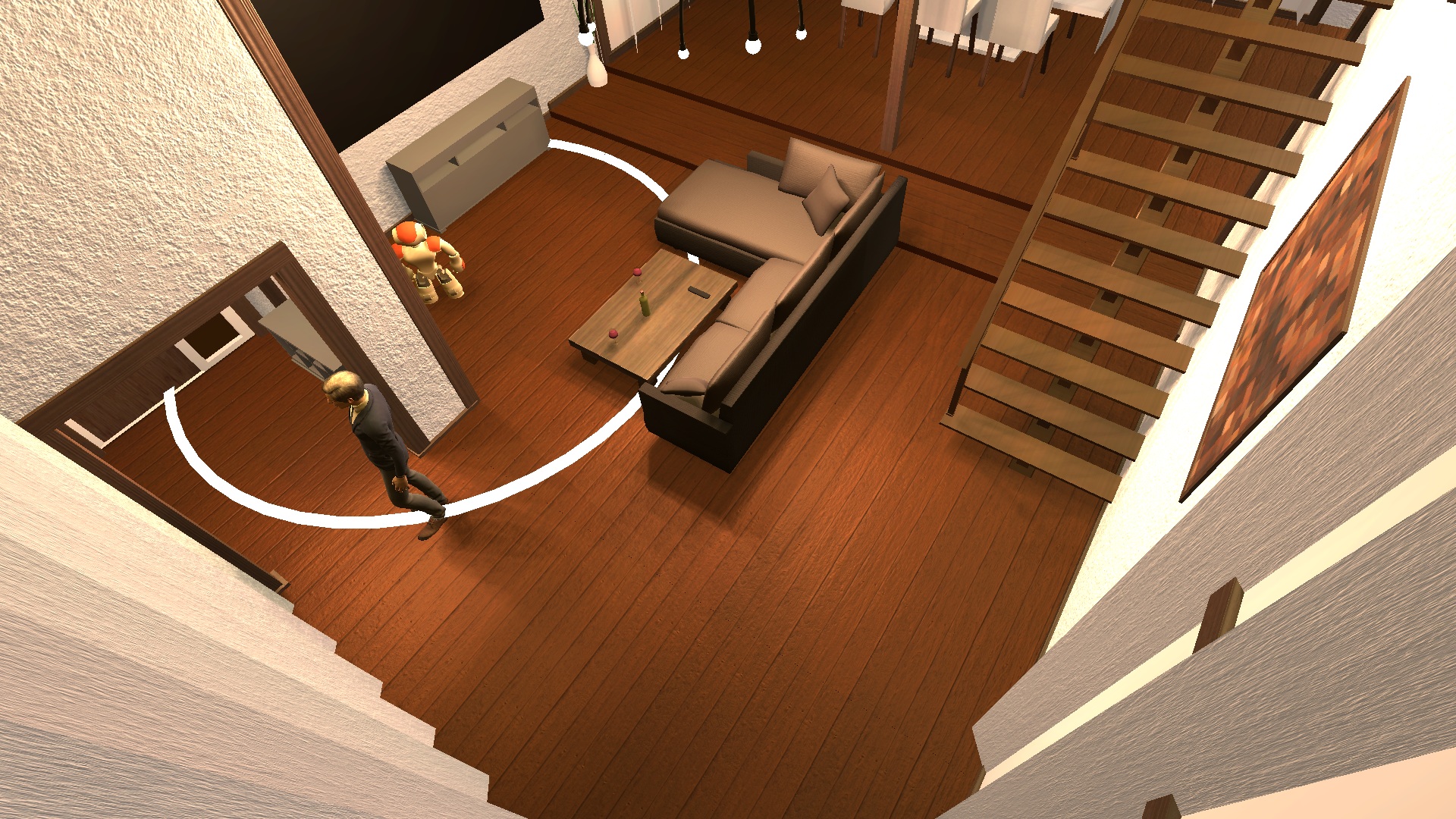}
    \end{subfigure}
    \begin{subfigure}[b]{0.321\linewidth}
        \centering
        \caption{\textbf{Scene B}}
        \includegraphics[width=\linewidth]{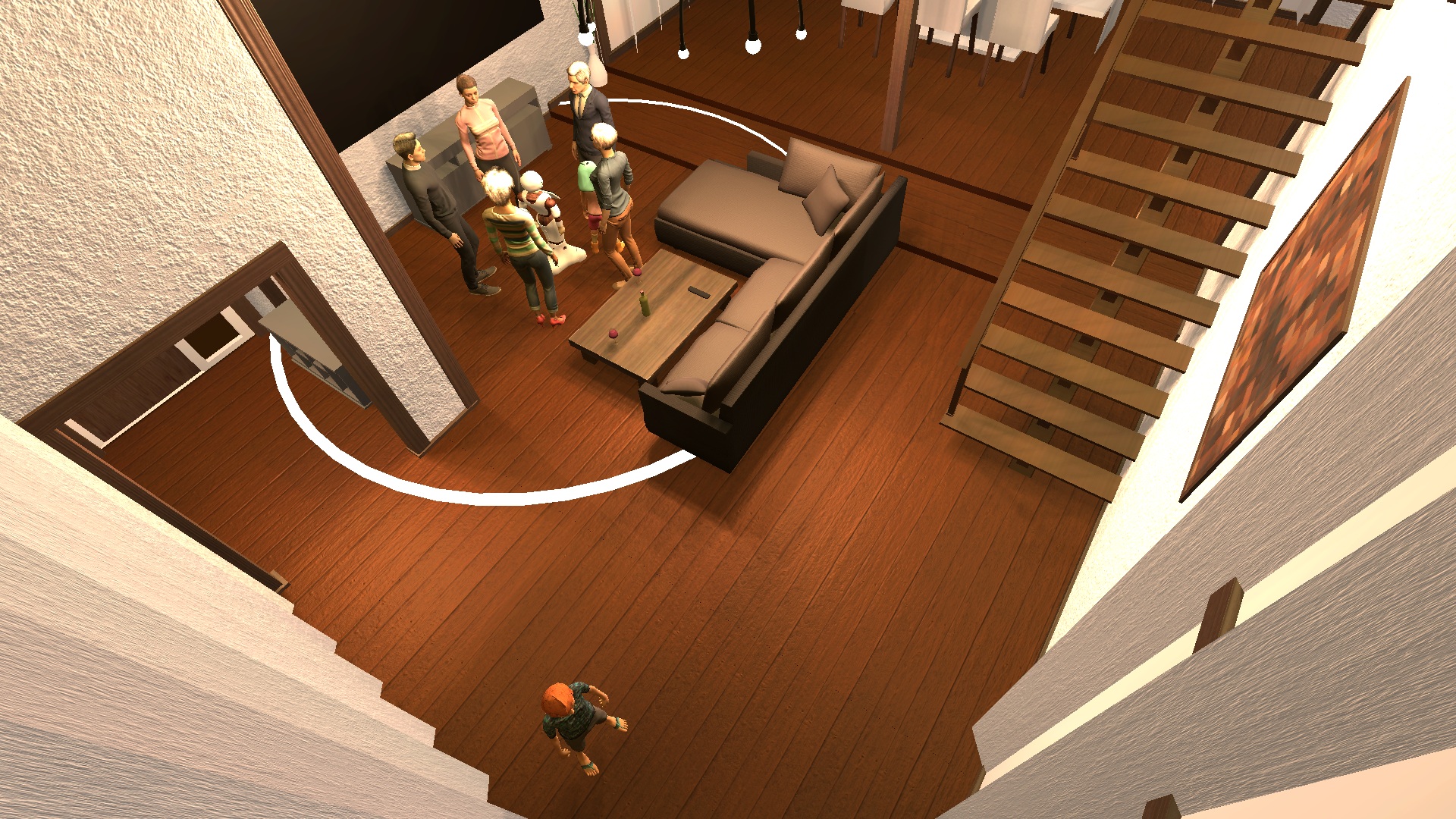}
    \end{subfigure}
    \begin{subfigure}[b]{0.321\linewidth}
        \centering
        \caption{\textbf{Scene C}}
        \includegraphics[width=\linewidth]{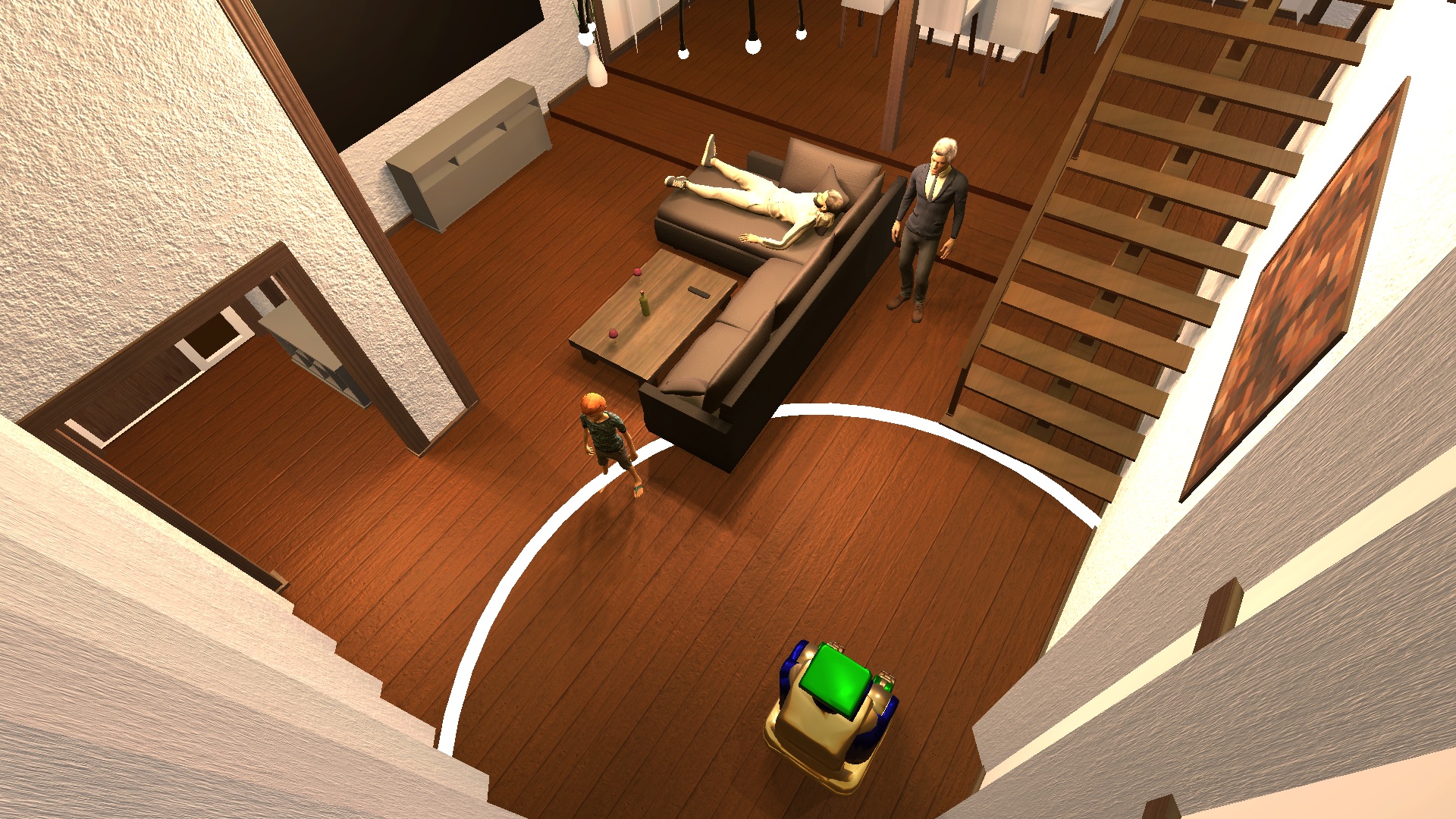}
    \end{subfigure}

    \vspace{0.5em} 

    \begin{tabular}{|c|c|c|c|c|c|c|c|c|c|}
    \hline
    \textbf{Scene} & \textbf{Vac.} & \textbf{Mop.} & \textbf{Warm F.} & \textbf{Cold F.} & \textbf{Drink} & \textbf{S. Obj.} & \textbf{L. Obj.} & \textbf{Clean.} & \textbf{Conv.} \\ \hline\hline
    \textbf{A} & 5 & 5 & 5 & 4 & 4 & 5 & 5 & 5 & 3 \\  
    \multicolumn{10}{|l|}{\textit{Expl.:} Safe ($p$: 0.82), scene not crowded ($p$: 0.80), robot far from others ($p$: 0.80)} \\\hline\hline
    \textbf{B} & 1 & 1 & 4 & 4 & 4 & 2 & 2 & 2 & 5\\  
    \multicolumn{10}{|l|}{\textit{Exp:} Robot close ($p$: 0.73), scene crowded ($p$: 0.72), robot facing people ($p$: 0.68)} \\ \hline\hline
    \textbf{C} & 2 & 5 & 4 & 4 & 3 & 4 & 4 & 2 & 3 \\  
    \multicolumn{10}{|l|}{\textit{Exp:} People not avail. ($p$: 0.72), robot capable ($p$: 0.62), robot not facing ($p$: 0.57)} \\ \hline
    \end{tabular}

    \caption{Given the human scores, the most likely explanations generated by the robot (prob. in parenthesis). The actions are \textit{ vacuum cleaning, mopping the floor, carrying warm food, carrying cold food, carrying drinks, carrying small objects, carrying large objects, cleaning, and starting a conversation}.}
    \label{fig:explanations}
    \vspace{-2.3em}
\end{figure}

In the previous section, we compared different LLMs to assess their accuracy while predicting the social appropriateness of robot actions. In this section, we evaluated the capability of the GRACE system to correct these LLM predictions to better match the human scores by leveraging human explanations. We present qualitative results where we compare GRACE with varying baselines based on their score prediction performances on test splits for condition $c_O$. We also present qualitative results where we analyze GRACE outputs based on generated robot explanations for condition $c_1$.\\
\textbf{Quantitative Results:} We include baselines that follow two different approaches for generating socially appropriate robot actions, as detailed in Section~\ref{Generative_training}. First, the left side of Table~\ref{combined_tables} shows the baselines that generated social appropriateness scores using the scene features. The results were obtained using all the MannersDB+ data (except one scene with insufficient annotations). 
These results show that GRACE outperforms the former methods by producing closer score values to the human ground truth, i.e., lower RMSE 
(STD: 0.01 both for MLP~\cite{tjomsland2022mind} and GRACE), and it captures the trend of human scores the best by providing the highest PCC and CCC values. 
The second set of baselines includes generative models that correct LLM predictions without leveraging human explanations to allow us to assess the impact of including explanations in the prediction process. These results are reported on the right side of Table~\ref{combined_tables} for MannersDB+ by discarding samples that did not include explanations belonging to the explanation categories described in Section~\ref{categorisation}. The results show that the proposed GRACE system performs the best based on average RMSE (STD: 0.01 for all models), PCC, and CCC, showing the significance of human explanations while generating socially appropriate robot actions. An extension of GRACE with a noising mask did not change the results noticeably.\\
\textbf{Qualitative Results:} To better understand the quantitative results of GRACE, we qualitatively assessed the explanations it generated as shown in Figure~\ref{fig:explanations}. The scenes were randomly selected from MannersDB+ and were not seen by the GRACE model during training. The results show that GRACE successfully outputs explanations and captures the scene features without explicitly being provided with scene attributes in its input space. 
For instance, based on human scores, GRACE successfully concludes that the robot's actions should be \textit{`safe'} in Scene A, where only one person is present. For Scene B, it infers that the scene is \textit{`crowded'}, as well as the robot is \textit{`too close'} and \textit{`facing people'}. For Scene C, where a person is lying on the sofa, the model suggests that the human state should be \textit{`unavailable'}.
All these demonstrate the success of the GRACE system in capturing the relationship between human explanations and their social appropriateness scores.

\section{Discussion and Conclusion}
\vspace{-0.5em}

In this work, we propose GRACE, a full system to generate socially appropriate robot actions leveraging LLMs and human explanations. GRACE first classifies the scenes based on their certainty, and for `certain' cases, it directly outputs LLM predictions. To handle `uncertain' scenes, GRACE improves the LLM predictions with a conditional autoencoder network structure, which includes human explanations in the learning process. Our evaluations showed that GRACE successfully learns the relationship between human scores and their explanations by outperforming several baselines and successfully generating sensible robot explanations.


Crucially, our work shows that leveraging human explanations while predicting the social appropriateness of robot actions boosts performance. As shown in Table~\ref{combined_tables}, GRACE not only reduced error values (RMSE) but also increased correlation scores (PCC and CCC). 
This is critical because robots should not predict just an average score but account for the trend of human preferences to ensure socially appropriate actions. 
For example, one user may prioritize safety, preferring the robot avoid carrying warm food in a crowd, while another may value comfort, wanting the robot to serve warm food at a gathering. 
The GRACE autoencoder captures these diverse preferences by integrating human explanations into the prediction process.

\red{In addition to generating socially appropriate robot actions, GRACE can provide sensible explanations for human-provided scores (Figure~\ref{fig:explanations})}. 
\red{These explanations include critical scene attributes, such as crowd size, the robot’s proximity to others, human states, and the safety of actions, despite the system not being explicitly trained on scene features.} 
These capabilities are crucial for HRI, enabling robots to provide reasoning in uncertain situations, boosting trust~\cite{siau2018building, edmonds2019tale}, and improving human-robot collaboration~\cite{9889368}.

\red{With the uncertainty classification, GRACE addresses the challenge of predicting inherent dataset uncertainty, where annotators gave differing scores for the same image. This task is difficult even for humans and similarly challenging for ML models, as shown in Table~\ref{tab:uncertainty_pred_combined}. Such misclassifications could potentially cause GRACE to directly output LLM predictions for uncertain cases or ask for explanations even in certain situations. Although the models struggled with the task, MannersDB results outperformed MannersDB+, likely due to its higher average annotators per scene (14.73 vs. 3.08), resulting in more reliable variance. Hence, increasing the number of annotations may improve uncertainty classification by better capturing human score agreement.}

%

Future work can extend GRACE  in several ways: i) The uncertainty classification performances can be improved using probabilistic models (i.e., Bayesian Neural Networks~\cite{mackay1992practical} \& Deep Ensemble Methods~\cite{lakshminarayanan2017simple}); and 
ii) GRACE can be deployed on a robot to assess the effectiveness and accuracy of real-time interactions, \red{i.e., the focus of our future work.}

\bibliographystyle{IEEEtran}
\bibliography{references}

\end{document}